# DEVELOPMENT OF A VOICE CONTROLLED ROBOTIC ARM

**Humayun Kabir[1], Akkas U. Haque[1*], S. C. Banik[2] and M. T. Islam[3]**

[1] Student, Department of Mechanical Engineering, CUET, Bangladesh
[2] Assistant Professor, Department of Mechanical Engineering, CUET, Bangladesh
[3] Professor, Department of Mechanical Engineering, CUET, Bangladesh
Chittagong University of Engineering & Technology, Chittagong-4349, Bangladesh
[1]rajume06@yahoo.com, *akkasuddin@gmail.com,[2]baniksajal2yahoo.com,[3]tazul2003@yahoo.com

This paper describes a robotic arm with 5 degrees-of-freedom (DOF) which is controlled by human voice and has been developed in the Mechatronics Laboratory, CUET. This robotic arm is interfaced with a PC by serial communication (RS-232). Users' voice command is captured by a microphone, and this voice is processed by software which is made by Microsoft visual studio. Then the specific signal (obtained by signal processing) is sent to control unit. The main control unit that is used in the robotic arm is a microcontroller whose model no. is PIC18f452. Then Control unit drives the actuators, (Hitec HS-422, HS-81) according to the signal or signals to give required motion of the robotic arm. At present the robotic arm can perform a set action like pick & pull, gripping, holding & releasing, and some other extra function like dance-like movement, and can turn according to the voice commands.

## 1. INTRODUCTION

Nowadays industries, service centers(Hospital), shopping centers, and house hold works are fully dependent on robotics & automation. For example, in a surgery, there are very few people besides the surgeon who actually contribute to the surgery. Most of the other people are there just to hand different tools and instruments to the surgeon, who is the one who actually does the surgery. Or, take the example of a mechanic. A mechanic almost always encounters situations in which he is forced to use a helper or assistant to do different things like holding two pieces of a machine together while welding, etc. It is thus seen that extra workforce is required, workforce that could otherwise be set to do other tasks. This is where the voice controlled robotic arm comes in. A robotic arm that is voice controlled enables the user to have more control over whatever task he is doing and also eliminates the need of unnecessary workforce. Also, the amount of stability and precision offered by the robot will be an added advantage over human assistants.

That is why, we think about a robot which will be controlled by human voice command that is versatile and can be used in a variety of different atmospheres and scenarios. We have successfully completed the first step in achieving our goal. That is, our robot can now listen to vocal commands given to it and respond accordingly. The final plan involves the robot understanding common phrases from natural speech and acquiring the ability to work seamlessly and in perfect coordination with a person.

## 2. THE ROBOTIC ARM

To investigate the feasibility of a robot that operates on the basis of natural speech processing, we built a robotic arm that responds to basic commands given to it.

### 2.1 Working

A microphone receives the voice commands and feeds them to the computer. The SAPI[1] engine detects the commands given and matches the commands with the dictionary created before. Once the commands are decoded, the necessary coordinates are fed to the function that calculates the angles required at each joint by the inverse kinematics mode of mechanics. These angles are then converted to the corresponding on-times required for the Pulse Width Modulation (PWM) for the servos. This information is then fed to the microcontroller via the USART mode of communication. The microcontroller used is an 8-

---

* Corresponding Author: Akkas U. Haque,
E-mail: akkasuddin@gmail.com
1. Speech Application Programming Interface (by Microsoft)

bit microcontroller, PIC16F852 from the microchip family.

The microcontroller then sends the required pulses to the servos attached to the each of the joints of the robot.

## 2.2 Flow of Control

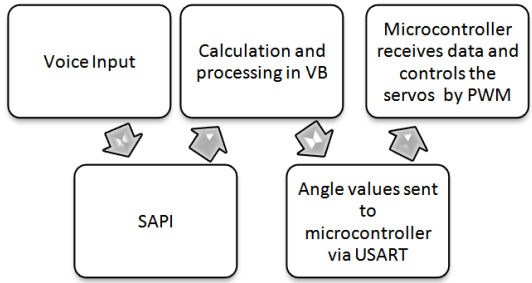

Fig. 1: Flow of control

## 2.3 Arm Overview
The robotic arm we used in this exploratory research is a small hobbyist device called the Lynx 6 by Lynxmotion. The arm has a total of five degrees of freedom (DOF): shoulder rotation, shoulder bend, elbow bend, wrist rotate, and wrist bend. A simple 2-prong gripper at the end of the arm is used to hold small objects. Figure 5 demonstrates these controllable features. Although the arm has no feedback or sensors, it is still sufficient as a prototype arm for use in this proof of concept exploration.

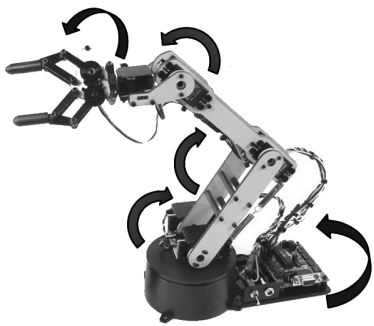

Fig. 2: Lynxmotion Lynx 6

The circuit board provided along with the package was replaced by one that we designed specifically for our purpose.

## 2.4 Speech Recognition
The Speech Application Programming Interface or SAPI is an API developed by Microsoft to allow the use of speech recognition and speech synthesis within Windows applications. The SAPI engine is incorporated as a part of the Visual basic program, developed by us, which is responsible for decoding the vocal commands and sending the coordinates to the microcontroller. The SAPI engine is also responsible for the computer responding by means of speech. The version used in our research is SAPI4.0 which has features that include Voice Command, Voice Dictation, Direct Speech Recognition, Direct Text To Speech etc.

## 2.5 Artificial Neural Network
An Artificial Neural Network (ANN) is an information processing paradigm that is inspired by the way biological nervous systems, such as the brain, process information. Artificial neural networks are made up of interconnecting artificial neurons (programming constructs that mimic the properties of biological neurons). Artificial neural networks may either be used to gain an understanding of biological neural networks, or for solving artificial intelligence problems without necessarily creating a model of a real biological system. The real, biological nervous system is highly complex and includes some features that may seem superfluous based on an understanding of artificial networks.

Every neural network possesses knowledge which is contained in the values of the connections weights. Modifying the knowledge stored in the network as a function of experience implies a learning rule for changing the values of the weights.

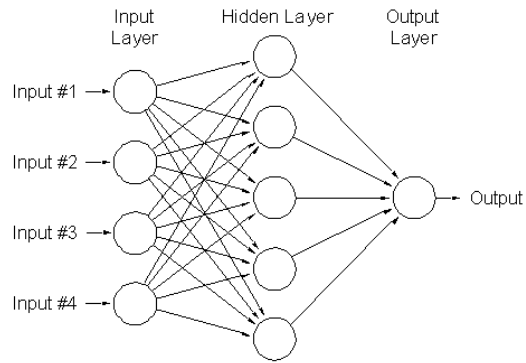

Fig. 3: Artificial Neural Network

Information is stored in the weight matrix W of a neural network. Learning is the determination of the weights. Following the way learning is performed, we can distinguish two major categories of neural networks:

**Fixed networks** in which the weights cannot be changed. In such networks, the weights are fixed a priori according to the problem to solve.

**Adaptive networks** which are able to change their weights.

All learning methods used for adaptive neural networks can be classified into two major categories:

**Supervised learning** which incorporates an external teacher, so that each output unit is told what its desired response to input signals ought to be. During the learning process global information may be required. Paradigms of supervised learning include error-correction learning, reinforcement learning and stochastic learning.

**Unsupervised learning** uses no external teacher and is based upon only local information. It is also referred to as self-organization, in the sense that it self-organizes data presented to the network and detects their emergent collective properties. Paradigms of unsupervised learning are Hebbian learning and competitive learning.

The SAPI engine uses the adaptive ANN of the first type i.e. Supervised Neural Learning. The computer uses the engine to collect a large amount of speech data from the user and trains itself to recognize the sounds that it receives and interpret the words correctly. The training generates a dictionary of words and phrases that are continually updated with more training. Once ample training is given to the SAPI engine, it is then able to recognize the words or commands given to it easily. After training the engine, we then wrote a program in Microsoft Visual Basic that uses this engine to communicate with the user. At present, our program can recognize around 100 different commands and respond vocally as well as do the work it was designed to do, i.e. control the robotic arm.

## 2.6 Serial Communication
We chose serial communication (RS-232) to interface our robotic arm with the computer. Serial communication was chosen because most of the computers have the necessary hardware and also because, it can be used over long distances without much loss in signal strength.

**Serial communication** is the process of sending data one bit at a time, sequentially, over a communication channel or computer bus. **RS-232** (Recommended Standard 232) is a standard for serial binary single-ended data and control signals connecting between a *DTE* (Data Terminal Equipment) and a *DCE* (Data Circuit-terminating Equipment). It is commonly used in computer serial ports. The standard defines the electrical characteristics and timing of signals, the meaning of signals, and the physical size and pin out of connectors. The signals of RS-232 serial are not suitable for use in TTL(**Transistor–transistor logic)** compatible digital logic circuits. So, **MAX232** integrated circuit is used to convert signals of **RS-232** to make suitable signals for use in TTL(**Transistor–transistor logic)** compatible digital logic circuits. The MAX232 is a dual driver/receiver and typically converts the RX, TX, CTS and RTS signals. It is helpful to understand what occurs to the voltage levels. When a MAX232 IC receives a TTL level to convert, it changes a TTL Logic 0 to between +3 and +15 V, and changes TTL Logic 1 to between -3 to -15 V, and vice versa for converting from RS232 to TTL. RS-232 Voltage Levels are given below:

**Table1: Voltage levels**

| RS232 Line Type & Logic Level | RS232 Voltage | TTL Voltage to/from MAX232 |
|---|---|---|
| Data Transmission (Rx/Tx) Logic 0 | +3 V to +15 V | 0V |
| Data Transmission (Rx/Tx) Logic 1 | -3 V to -15 V | 5V |
| Control Signals (RTS/CTS/DTR/DSR) Logic 0 | -3 V to -15 V | 5V |
| Control Signals (RTS/CTS/DTR/DSR) Logic 1 | +3 V to +15 V | 0V |

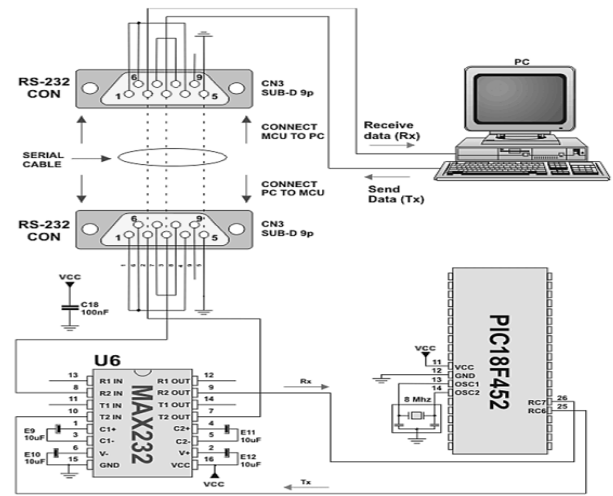

Fig. 4: Serial Communication(PC- microcontroller)

## 2.7 Pulse-Width Modulation

Pulse-width modulation (PWM) is a commonly used technique for controlling power to inertial electrical devices, made practical by modern electronic power switches. We used PWM to control Servo motor (HS-422 & HS81). PWM, as the name suggests, is a method of controlling devices by varying the length of the pulse in a given time period.

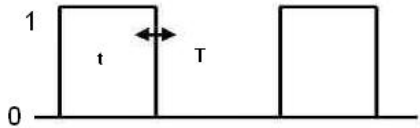

Fig. 5: Duty cycle

Duty Cycle describes the proportion of 'on' time to the regular interval or 'period' of time; a low duty cycle corresponds to low power, because the power is off for most of the time. Duty cycle is expressed in percent, 100% being fully on.

Duty Cycle, $D = \dfrac{t}{T}$

Where,
t = on state or high state
T = the period of the function.

We chose the servos because they have the control circuits built in, and they have the highest torque to weight ratio and also because they can be precisely controlled by PWM. On the other hand, stepper motors and DC motors were both out of question due to weight constraints.
In Servo Motor Control, the time period of oscillation is 20ms. And the on-time varies from 1ms for 0 degrees to 2ms for 180 degrees.

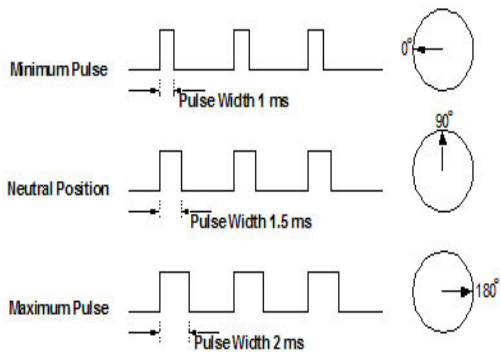

Fig. 6: Servomotor control PWM diagram

## 2.8 Inverse Kinematics

The kinematics solution of any robot manipulator consists of two sub problems, forward and inverse kinematics. Forward kinematics determines where the robot's manipulator hand will be if all joints are known whereas inverse kinematics is used to find out where each joint variable must be if the desired position and orientation of end-effector is pre-determined.

The kinematics that we applied in this research was Inverse Kinematics. As shown in the diagram, the x and y coordinates of the end effector on the plane along a specified angle of the base is known. The angles at each joint are then calculated based on the equations derived by IK method.

### 2.8.1 Geometric Solution for IK Equations of the Arm

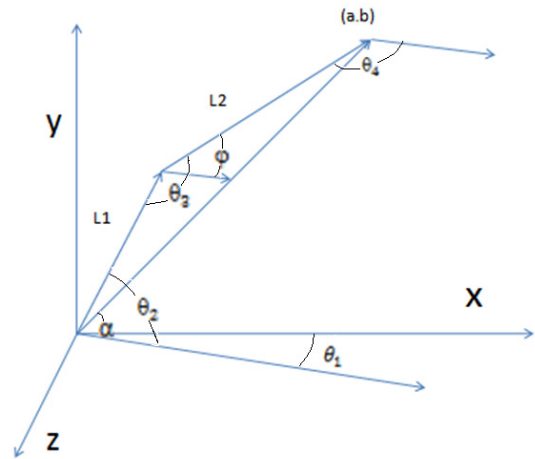

Fig. 7: Geometric Solution for IK equations

Let the coordinates of the end effector be (a,b, $\theta_1$) where a,b are the coordinates of the effector in the plane which is offset at an angle of $\theta_1$ from the xy plane.

From the figure,

$$a = l_1 \cos\theta_2 + l_2 \cos\varphi$$
$$b = l_1 \sin\theta_2 + l_2 \sin\varphi$$

Rearranging, we get,

$$\cos(\theta_2 - \varphi) = \dfrac{a^2 + b^2 - (l_1^2 + l_2^2)}{2 l_1 l_2}$$

Taking $\gamma = \theta_2 - \varphi$, we get,

$$\gamma = \arccos\left(\dfrac{a^2 + b^2 - (l_1^2 + l_2^2)}{2 l_1 l_2}\right)$$

Taking,

$$\alpha = \arctan\left(\dfrac{b}{a}\right)$$

Now applying sine rule for the triangle OAB, we get,

$$\frac{\sin(\theta_2 - \alpha)}{l_2} = \frac{\sin(\alpha - \varphi)}{l_1} = \frac{\sin(\theta_3)}{\sqrt{a^2 + b^2}}$$

Solving we get the values of $\theta_2, \theta_3$ and $\theta_4$ as,

$$\theta_2 = \alpha + \arcsin\left(\frac{l_2 \sin \gamma}{\sqrt{a^2 + b^2}}\right)$$
$$\theta_3 = \pi - \gamma$$
$$\theta_4 = \pi + \arcsin\left(\frac{l_1 \sin \gamma}{\sqrt{a^2 + b^2}}\right) - \alpha$$

## 3. CONCLUSIONS

The robotic arm was tested with different users and with sufficient training was able to respond to commands with an accuracy of almost 90 percent. Based on our study, we were able to conclude that it is possible to create robots for industrial applications that could interact with humans verbally and also help users to do their required tasks quickly and efficiently. At present our robotic Arm can perform griping, holding & releasing object and dancing according to user voice command. It can also perform the conversation with users. This robotic can be developed without PC interfacing by using DSP(digital signal processing) module or HM-2007 & SPO-256 IC's.